\documentclass[]{article}
\usepackage[letterpaper]{geometry}
\usepackage{mtsummit2021}
\usepackage{times}
\usepackage{url}
\usepackage{latexsym}
\usepackage{natbib}
\usepackage{layout}
\usepackage{pgfplots}
\usepackage{multirow}
\usepackage{amsmath}
\usepackage{booktabs}

\newcommand{\eob}{\texttt{$<$eob$>\:$}}
\newcommand{\eol}{\texttt{$<$eol$>\:$}}
\newcommand{\eos}{\texttt{$<$eos$>\:$}}
\newcommand\blfootnote[1]{%
  \begingroup
  \renewcommand\thefootnote{}\footnote{#1}%
  \addtocounter{footnote}{-1}%
  \endgroup
}

\usepackage{xcolor, colortbl}
\usepackage{wasysym}

\begin{document}

\title{\bf Simultaneous Speech Translation for Live Subtitling: from Delay to Display  }
\author{\name{\bf Alina Karakanta $^{*\eighthnote\halfnote}$} \hfill  \addr{akarakanta@fbk.eu}\\
        \name{\bf Sara Papi $^{*\eighthnote\halfnote}$} \hfill \addr{spapi@fbk.eu}\\
        \name{\bf Matteo Negri $^{\eighthnote}$} \hfill \addr{negri@fbk.eu}\\
       \name{\bf Marco Turchi $^{\eighthnote}$} \hfill \addr{turchi@fbk.eu}\\
        \addr{$^{\eighthnote}$ Fondazione Bruno Kessler, Via Sommarive 18, Povo, Trento, Italy  \\ $^{\halfnote}$ University of Trento, Italy}\vspace{-.25cm}
}

\maketitle
\pagestyle{empty}
\vspace{-.25cm}
\begin{abstract}
With the increased audiovisualisation of communication, the need for live subtitles in multilingual events is more relevant than ever. In an attempt to automatise the process, we aim at exploring the feasibility of simultaneous speech translation (SimulST) for live subtitling. However, the word-for-word rate of generation of SimulST systems is not optimal for displaying the subtitles in a comprehensible and readable way
. In this work, we adapt SimulST systems to predict subtitle breaks along with the translation. We then propose a display mode that exploits the predicted break structure by presenting the subtitles in scrolling lines. We compare our proposed mode with a display 1) word-for-word and 2) in blocks, in terms of reading speed and delay. Experiments on three language pairs (en$\rightarrow$it, de, fr) show that scrolling lines is the only mode achieving an acceptable reading speed while keeping delay close to a 4-second threshold. We argue that simultaneous translation for readable live subtitles still faces challenges, the main one being poor translation quality, and propose directions for steering future research.
\end{abstract}

\section{Introduction}\blfootnote{$^*$Equal contribution}
The globalisation of business, education and entertainment, together with the recent movement restrictions, have 
transferred human interaction to the online sphere. The boom in online multilingual communication is setting new challenges for achieving barrierless interaction between audiences with diverse linguistic and accessibility needs. Subtitles, as a key means for ensuring accessibility, have been adapted to respond to the challenge of timely communication, giving rise to live subtitling. Live subtitling, whether intralingual (same language as the speech) or interlingual (different language than the speech), allows for obtaining subtitles in real time and is recently witnessing an upsurging demand in a range of occasions; from news and TV programmes, to online meetings, conferences, live events, shows and university lectures, live subtitles are bringing the world closer together. 

Technology has always been a leading factor in live subtitling. 
Live subtitles were initially obtained by means of keyboards or stenotyping, but Automatic Speech Recognition (ASR) gave rise to respeaking, a technique where the subtitler respeaks the original sound into an ASR system, which produces the subtitles on the screen \citep{romero-fresco-2011-respeaking}. Although originally employed for intralingual subtitles, respeaking is gradually extending to produce interlingual subtitles, a task bearing close resemblance to simultaneous interpreting. Still, the tediousness of the task and the scarcity of highly-skilled professionals for live subtitling call for a more pronounced role of technology for providing real-time access to information. 

These growing needs for access to multilingual spoken content have motivated researchers to develop fully automatic solutions for real-time spoken language translation \citep{grissom-ii-etal-2014-dont,gu-etal-2017-learning,alinejad-etal-2018-prediction,arivazhagan-etal-2019-monotonic,ma-etal-2019-stacl}. The new possibilities opened up by 
neural machine translation have led to improvements in automatic 
simultaneous speech-to-text translation (SimulST). 
In SimulST \citep{ma-etal-2020-simulmt,ren-etal-2020-simulspeech}, the generation of the translation starts before the entire audio input is received, which is an indispensable characteristic for achieving low latency (translation delay) between speech and text in live events. The translation becomes available at consecutive steps, usually one word at a time. However, a display mode based on the word-for-word rate of generation of SimulST systems may not be optimal for displaying readable subtitles. Studies in intralingual subtitling have shown that a word-for-word display increases the number of saccadic crossovers between text and scene \citep{Rajendran-et-al-2013} and leaves viewers less time to look at the images \citep{Romero-fresco-2010-Standingonquicksand}. For this reason, regulators, such as the UK Office of Communications, recommended displaying subtitles in blocks \citep{ofcom-2015}. However, this display mode is not ideal for live events since waiting until the block is filled before displaying the subtitle would extremely increase latency at the risk of losing synchronisation with the speaker. Despite the existence of some applications of SimulST, so far no work has explored its potential for live subtitling and how the delay in generation impacts the readability of the subtitles. 

Given the boosting demand in live subtitles and previous studies on the readability of live subtitles, in this work we pose the following research questions: \textbf{1) Can automatic simultaneous translation be a viable method for producing live interlingual subtitles? 2) What are the challenges of the generation mode of SimulST systems for the readability of the subtitles?}
We first explore the performance of a direct SimulST system on three language pairs (en$\rightarrow$it, de, fr) in terms of translation quality and its ability of generating readable subtitles in terms of technical constraints, such as length and proper segmentation. Second, we investigate two methods for displaying live subtitles,
 i.e. i) word-for-word and ii) blocks, and how the display mode affects their readability (reading speed and delay). Thirdly, we propose scrolling lines, a mixed display method which takes advantage of the ability of our system to define proper line breaks and show that it leads to a more comfortable reading speed at an acceptable delay. Lastly, we discuss challenges and recommendations for applying simultaneous translation for live subtitling.

\section{Related work}

\subsection{Live subtitling and its reception}
Live subtitles 
are a simpler and more customisable alternative to other ways of translating speech, such as simultaneous interpreting \citep{marsh-2004}.
Originally, the practice of live subtitling was used to produce intralingual subtitles, 
in order to enable deaf or hard-of-hearing persons to follow live TV programs \citep{lambourne-2006}. With the widening definition of accessibility beyond the deaf and hard-of-hearing to include persons not speaking the source speech language
, live subtitling was adapted to provide interlingual subtitles \citep{Dawson_2019}.

Live subtitles were produced initially with standard keyboards, but the need to reduce latency led to resorting to stenography or to the invention of a customised syllabic keyboard, called Velotype\footnote{\url{https://www.velotype.com/en/homepage-eng/}}. With the adoption of ASR technologies, respeaking became the most popular technique for live subtitling \citep{lambourne-2006}. With this technique, a respeaker listens to the original sound of a (live) event and respeaks it to an ASR software, which turns the recognized utterances into subtitles (Romero-Fresco 2011).
In its interlingual mode, live subtitling is a newly established practice and therefore the industry is experimenting with different profiles for the role of interlingual live subtitlers \citep{LANS-TTS515}. Interlingual live subtitling requires skills from three disciplines: respeaking, subtitling and simultaneous interpreting. As a result, the availability of highly-skilled professionals for interlingual live subtitling cannot meet the growing needs in real-time multilingual communication.

Except for the quality of live subtitles, their speed and display mode greatly affect the user's views, perception and comprehension \citep{perego-et-al-2010}. The faster the subtitles, the more time users spend on reading them, and therefore they have less time to focus on the images, which negatively impacts comprehension. 
Recommendations for comfortable reading speed depend on the user group and language. For example, 15 characters per second (cps) are recommended for live interlingual English SDH -- subtitles for the deaf and hard of hearing -- \citep{ofcom-2005}, 12-15 cps for offline interlingual subtitling in Central Europe \citep{Szarkowska-2016-report}, 17–20 cps in global online streaming services \citep{Netflix} and 21 cps for TED Talks \citep{ted}. According to \cite{Romero-Fresco-2015-speed}, a fast subtitle speed of 17–18 cps allows viewers to spend approximately 80\% time on subtitles and only 20\% on images. 
As for the display mode, \citet{Romero-fresco-2010-Standingonquicksand} found that a word-for-word display results in viewers spending 90\% of time reading the subtitles as opposed to 10\% looking at the images, which detriments comprehension. Moreover, the presence of the word to the right of fixation is vital for fluent reading \citep{Rayner-et-al-2006-right} and its absence leads to more re-reading \citep{Sharmin-et-al-2016-readingdynamic}. \citet{Rajendran-et-al-2013} showed that scrolling subtitles cause the viewers to spend significantly more time reading than subtitles appearing in blocks. These findings have been assumed by broadcasters in several countries to replace their scrolling subtitles by block subtitles where possible. Currently, a word-for-word display is used in most live speech translation applications, such as STACL \citep{ma-etal-2019-stacl}, ELITR \citep{bojar-etal-2021-elitr} and Google Translate \citep{RetranslationVS}. In this work, we experiment with displaying the output of SimulST systems in ways which turn out to be more comfortable for the viewer, leading to better comprehension and a more pleasant user experience.

\subsection{Simultaneous translation}
Simultaneous Speech Translation (SimulST) is the task in which the generation of the translation starts before the audio input becomes entirely available.
In simultaneous settings, a model has to choose, at each time step, a read or a write action, that is, whether to receive new information from the input or to write using the information received until that step. Consequently, a SimulST system needs a policy which decides the next action.
Decision policies can be divided into: \emph{fixed}, when the decision is taken based on the elapsed time, and \emph{adaptive}, when the decision is taken by looking also at the contextual information extracted from the input. Even if the adoption of a fixed policy disregards the input context leading to a sub-optimal solution, little research has been done on adaptive policies \citep{gu-etal-2017-learning,zheng-etal-2019-simpler,Zheng2020SimultaneousTP} because they are hard and time-consuming to train \citep{zheng-etal-2019-simultaneous,arivazhagan-etal-2019-monotonic}. 

Among the fixed policies, the most popular and recently studied is the wait-$k$ strategy, which was first proposed by \cite{ma-etal-2019-stacl} for simultaneous Machine Translation (SimulMT).
The SimulMT wait-$k$ policy is based on waiting for $k$ source words before starting to generate the target sentence.
This simple yet effective approach was then employed in SimulST, as in \cite{ma-etal-2020-simulmt} and \cite{ren-etal-2020-simulspeech}, by using direct models i.e. models that, given an audio source, generate a textual target without the need for intermediate transcription steps.

While the original wait-k implementation is based on textual source data, \cite{ma-etal-2020-simulmt} adapted this strategy to the audio domain by waiting for $k$ fixed amount of time (step size) instead of $k$ words. The best step size resulting from their experiments was 280ms, corresponding to, approximately, the length of a word -- on average 271ms -- motivating the equivalence between the MT and the ST policies. 
In \cite{ren-etal-2020-simulspeech}, the adaptation was done differently since their direct system includes a segmentation module that is able to determine word boundaries i.e. when a word finishes and the successive one starts. In this case, the wait-k strategy is applied by waiting for $k$ pauses which are automatically detected by the segmenter.

Some studies have attempted to improve the performance of the wait-$k$ strategy, both in relation to latency and quality. For instance, \cite{nguyen2021empirical} propose to emit more than one token during the writing mode to improve the quality-latency trade-off, while \cite{Elbayad2020} propose a unidirectional encoder instead of a standard SimulST bidirectional encoder (i.e. avoiding to update the encoder states after each READ action) to slow down the decoding phase. However, these systems are not applicable in our case since \cite{nguyen2021empirical} uses an offline system which is simulated as a simultaneous system during the decoding phase while the model of \cite{Elbayad2020} is for SimulMT and not for SimulST.
No previous work has explored the possibilities offered by SimulST for the generation of live subtitles.

\section{SimulST for live subtitling}

\subsection{Simultaneous ST models}

The SimulST systems used in this work are based on direct ST models \citep{berard_2016,weiss2017sequence}, which are composed of an audio encoder and a text decoder. The encoder starts from the audio features extracted from the input signal and computes a hidden representation, while the decoder transforms this representation into the target text.
These systems have been shown to have 
lower latency \citep{ren-etal-2020-simulspeech} -- an important factor in simultaneous systems -- compared to cascade systems, which perform two generation steps, one for transcription and one for translation. 
Moreover, \cite{karakanta-etal-2020-42} suggested that direct ST systems, having access to the audio source, make better subtitle segmentation decisions by taking advantage of the pauses in the audio. 

In order to adapt SimulST systems for the task of live interlingual subtitling, we force the system to learn from human subtitle segmentation decisions by training on data annotated with break symbols which correspond to subtitle breaks (\eob for end of a subtitle block and \eol for end of line inside a subtitle block). These break symbols, if positioned properly, are the key element which allows us to experiment with different display modes for live subtitles.

Our direct SimulST models combine the efficiency of the wait-$k$ strategy \citep{ma-etal-2019-stacl} 
and the findings of \cite{karakanta-etal-2020-42} for obtaining readable subtitles with direct ST systems.
This decision policy was also chosen because it 
allows us to control the latency of our systems. In this way, we can study the effect of latency both on the conformity of the subtitles and on the subtitle display modes.

\subsection{Display modes}
\label{ssec:display}
\begin{figure*}[t!]
    \centering
    \includegraphics[width=\textwidth,height=8.4cm]{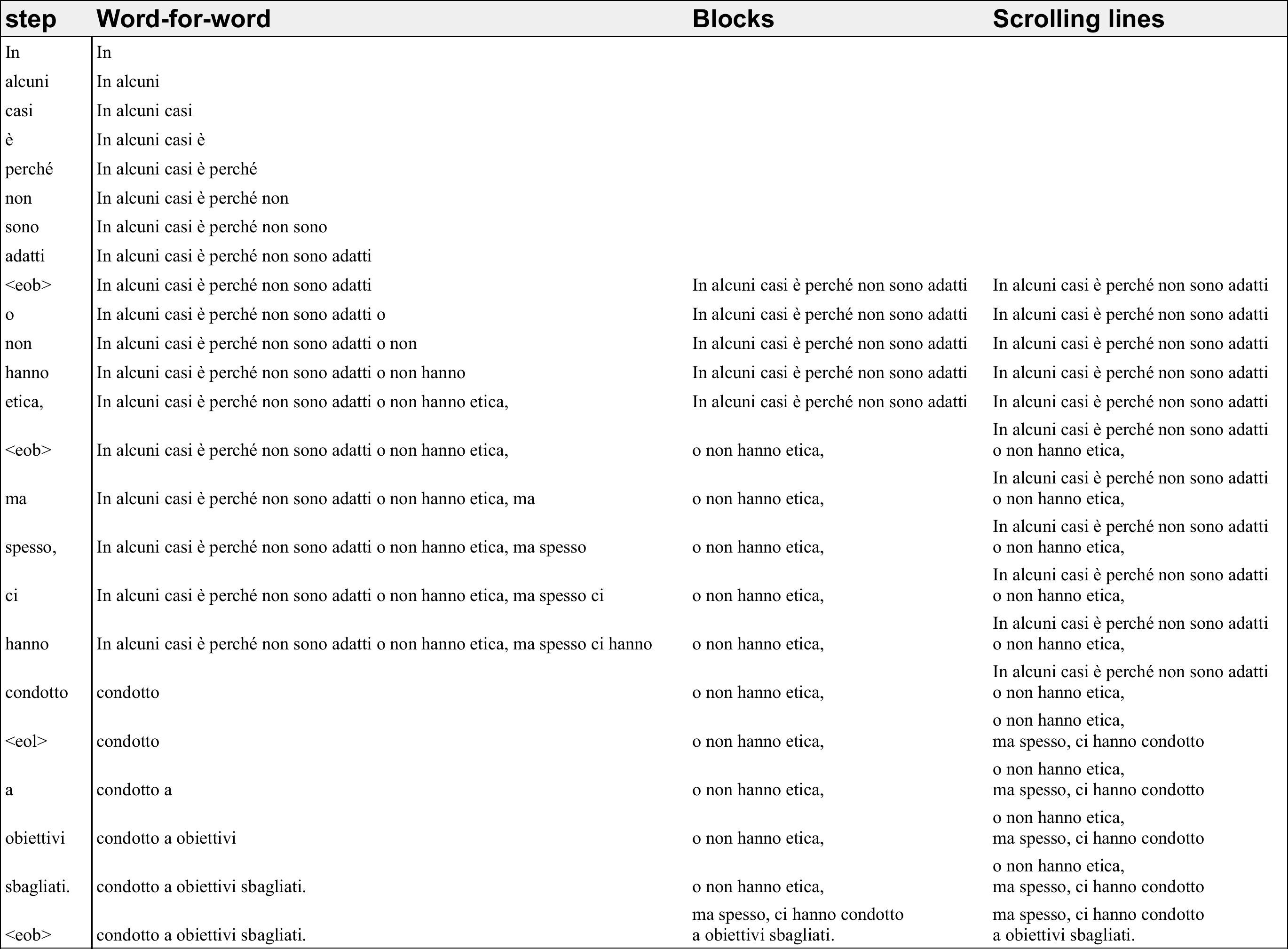}
    \caption{Example sentence displayed in the three different modes. Words on the left column correspond to the time steps.}
    \label{fig:example}
\end{figure*}

We experiment with the following three display modes: 1) word-for-word, 2) blocks, and 3) scrolling lines. Figure~\ref{fig:example} shows an example of a subtitle displayed in the different modes. 

\paragraph{Word-for-word:}\label{subsec:mode1} In the word-for-word display, words appear sequentially on the screen as soon as they are generated by the system. One line at the bottom of the screen is filled from left to right until no more space is available on the right. Then, the line disappears and a new line is filled again. As already mentioned, this display mode is used in most simultaneous applications and follows the generation process of the SimulST system. Naturally, the length of the subtitle depends on the size of the screen. In our case, we selected a maximum of 84 characters, which matches the max. length of a full subtitle block of the block method (see below) \footnote{According to the TED Talk subtitling guidelines.} and approximates the length observed in the STACL demo.\footnote{\url{https://simultrans-demo.github.io/}}

\paragraph{Blocks:}\label{subsec:mode2} In this mode, the subtitles are displayed only when a full subtitle block is completed. This means that the system continues generating words which are only displayed when the block delimiter \eob is generated. The subtitle block remains on screen until the next \eob symbol is generated, therefore the first subtitle is substituted by the next one. Display in blocks is used primarily in offline subtitling, where subtitles are prepared beforehand.

\paragraph{Scrolling lines:}\label{subsec:mode3} 
Instead of waiting for the full block, we propose a mode in which each line 
is displayed as soon as a break is predicted (either \eob or \eol). Whenever a new break is predicted, the previous line moves to the upper row of the block and the new line occupies the lower row. Since the allowed number of lines in a block is two, each line moves from the lower to the upper row before disappearing. This mode combines the benefits of the two previous methods. It reduces the dynamicity of the text compared to the word-for-word display, it makes content available earlier than the block display, it allows for access to extended context and it reduces long-distance eye movements since the previous line appears above. 
This mode is similar to the most popular mode employed for broadcasting, with the difference that there is no word-for-word display inside the lines.

The last two display modes are possible because of the ability of our SimulST system to predict subtitle breaks, which was not considered before in SimulST.

\subsection{Evaluation of display modes}
\label{ssec:eval}
Our evaluation of the generated subtitles follows Ofcom's recommendations \citep{ofcom-2015} and focuses on three key dimensions: quality, delay between utterance and subtitle, and reading speed. For the display modes the quality is fixed, since they are applied to the same output. We thus focus on the speed and delay. In general, the reading speed is computed in characters per second (cps), as the number of characters over the total time of display. A low but also constant reading speed is pivotal for user experience
, considering that it represents how fast or slow a user has to read.
Since each of the display methods described in Section~\ref{ssec:display} has a different granularity, the computation of the reading speed has to take into account these visualization differences. The computations of reading speed and delay are described in detail for each visualization mode below.\footnote{The code is available at: \url{https://github.com/sarapapi/reading_speed}}



\paragraph{Reading Speed.} In the \textbf{Word-for-word} mode (see Figure~\ref{fig:example}) a word appears on screen as soon as it is generated and remains on screen until the block changes, i.e. when the 84-character limit is reached. 
Thus, the amount of time available for a user to read this word and all the following words of the block -- hereinafter display time -- is the interval elapsed from the generation of the word to that of the first word of the successive block.
Consequently, the reading speed can be computed at each generation step as the length (in characters) of the generated word and of the successive words of the block divided by the display time.
After computing the reading speed for each word of a block, the block-level reading speed is obtained by taking the maximum value since this represents how fast a user has to read to avoid losing part of the text. 

Formally, each block corresponds to a group of words of maximum 84 characters or to the last group of words before end of sentence (\eos) is emitted, i.e. the end of the audio segment.
Thus, a block is composed by a set of $W$ words $w_1,...,w_{W}$ emitted at times $t_1,...,t_{W}$, measured in seconds. At time $t_{W+1}$, the successive block starts or \eos is emitted.

With this notation, we can compute the reading speed as follows:
\begin{equation}
    rs = \max_{i=1,...,W} \frac{len(text_i)}{elapsed_i} 
\end{equation}
where: \vspace{-.5cm}
\begin{equation}
\label{eq:elapsed}
    elapsed_i = \begin{cases} 
                \texttt{DELAY\_K}, & w_i = \eos \\
                t_{i+1}-t_i, & i=W \\ 
                t_{i+1}-t_i+elapsed_{i+1}, & otherwise
                \end{cases}
\end{equation}
\begin{equation}
\label{eq:text}
    text_i = \begin{cases} 
            w_i & i=W \\ 
            w_i+\text{\texttt{SPACE}}+text_{i+1} & otherwise
            \end{cases}
\end{equation}
and \texttt{SPACE} represents a blank space added between the two texts. If the block is the last block of the audio segment, we do not know the emission time of the next word. For this reason, we use a fictitious delay:
\begin{equation*}
    \texttt{DELAY\_K} = 0.280s \cdot k
\end{equation*}
where $k$ corresponds to the $k$ of the wait-k policy. This amount of time is a lower bound for the generation time of the first token of the next segment.
As the wait-k policy reads for $k$ steps (each of them lasting 280ms) and then generates the first word, the actual elapsed time will always be higher as it includes the time required for the generation of the word.
Thus, \texttt{DELAY\_K} represents a conservative estimation of the time available to read the last word.


For the \textbf{Blocks} mode, the block/line structure of the subtitles is exploited and a block stays on screen until the next block is filled. The reading speed is computed at block level, dividing its length by the time elapsed between the display time of the current block and that of the successive one. 
In an audio segment with $B$ blocks, each block $b$ is composed by $W_{b}$ words $w_1,..,w_{W_b}$, where $w_{W_b}$ is the \eob.
Each word $w_i$ is emitted at time $t_i$, thus each block $b$ is emitted at time $t_{W_b}$, hereinafter $t_b$.
Using this notation, Equations \ref{eq:elapsed} and \ref{eq:text} become:
\begin{equation}
\label{eq:elapsed_blocks}
    elapsed_b= \begin{cases} 
                \texttt{DELAY\_K}, & b=B     \\
                t_{b+1}-t_{b}, & otherwise
                \end{cases}
\end{equation}
\begin{equation}
\label{eq:text_blocks}
    text_b = \sum\limits_{i=1}^{W_b-1}
    \begin{cases} 
                \texttt{BLANK}, & w_i = \eol\\
                w_i, & i = W_b-1\\
                 w_i+\text{\texttt{SPACE}}, & otherwise
    \end{cases}
\end{equation}
where \texttt{BLANK} corresponds to the empty string and, as before, \texttt{DELAY\_K} conservatively accounts for the last unknown block time. In $text_b$ we do not consider \eol and \eob (the $W_b$-th word) for the reading speed computation since they are only used for formatting the subtitles and are not read by the user. Consequently, the reading speed of a block is:
\begin{equation}
\label{eq:rs_blocks}
    rs = \frac{len(text_b)}{elapsed_b}
\end{equation}

Our proposed display mode, \textbf{Scrolling Lines}, also exploits the subtitle structure considering both \eob and \eol as a unique \eol delimiter.
Since each line scrolls up when another is generated and two lines stay together on the screen, each line is displayed until the next two lines are generated. As a consequence, the reading speed is computed at line level, dividing the length of a line by the time needed to generate the two successive lines. 

If we denote $L$ as the number of lines present in an audio segment, then each line $l$ is composed by $W_{l}$ words $w_1,..,w_{W_l}$ emitted at times $t_1,...,t_{W_l}$, where $w_{W_l}$ is the \eol and $t_{W_l} = t_l$ is its emission time.
In this case, the reading speed is calculated at line-level instead of block-level, considering that each line is displayed until the next two lines (since a block can be composed by two lines) are produced. Thus, Equations \ref{eq:elapsed_blocks}, \ref{eq:text_blocks} and \ref{eq:rs_blocks} are modified as follows:
\begin{equation}
    rs = \frac{len(text_l)}{elapsed_l}
\end{equation}
\begin{equation}
    elapsed_l= \begin{cases} 
                \texttt{DELAY\_K}, & l=L     \\
                (t_{l+2}-t_{l+1}) + (t_{l+1}-t_{l}), & otherwise
                \end{cases}
\end{equation}
\begin{equation}
    text_l = \sum\limits_{i=1}^{W_l-1}
    \begin{cases} 
                w_i, & i = W_l-1\\
                 w_i+\text{\texttt{SPACE}}, & otherwise
    \end{cases}
\end{equation}
where, in this case, \texttt{DELAY\_K} conservatively accounts for the last unknown line time.

\paragraph{Delay.} The delay is estimated as the time between speech and subtitling. While in intralingual subtitling the correspondence between audio and subtitle is easier to establish, the interlingual setting poses the challenge of finding the correspondences between source audio and target text. Since a word-aligner would capture semantic correspondences, we opt for a temporal-based correspondence, based on the system's lagging. Therefore delay is calculated as:
\begin{equation}
delay = \sum\limits_{i=1}^{w_i} (t\_display_{w} - t\_received_{w} - \texttt{DELAY\_K})
\end{equation}

where $t\_display_{w}$ is the time the word was displayed on screen and $t\_received_{w}$ the utterance time corresponding to the displayed token, minus the $k$-wait delay. For word-for-word display, the time of display corresponds to the system's elapsed time, therefore the delay equals the system's lagging. For block and scrolling lines display, the time of display is the time elapsed for the generation of the break (\eob for blocks or any break for scrolling lines).

\section{Experimental setting}
\label{sec:expset}
\paragraph{Data} For our experiments we use MuST-Cinema \citep{karakanta-etal-2020-must}, an ST corpus compiled from TED Talk subtitles. This corpus is ideal for exploring display modes other than word-for-word because it contains subtitle breaks as special symbols. We conduct experiments on three language pairs, English$\rightarrow$Italian (442 hours), English$\rightarrow$German (408 hours) and English$\rightarrow$French (492 hours). For tuning and evaluation we use the MuST-Cinema dev and test sets.
The text data were tokenized using SentencePiece \citep{sentencepiece} with the unigram setting \citep{unigram}, trained on the training data with a 10k-token vocabulary size. 
The source audio was pre-processed with the SpecAugment data augmentation technique \citep{Park2019}, then the speech features (80 log Mel-filter banks) were extracted and Cepstral Mean and Variance Normalization was applied at global level. Samples with a length above 30s were filtered out.
The configuration parameters are the default ones set by \cite{ma-etal-2020-simulmt}. 

\paragraph{Training settings} Our SimulST systems are Transformer-based models \citep{Transformer}, composed by 12 encoder layers, 6 decoder layers, 256 features for the attention layers and 2,048 hidden units in the feed-forward layers. 
All models are based on a custom version of \cite{fairseq_s2t}, having two initial 1D convolutional layers with \textit{gelu} activation functions \citep{gelu}, but adapted to the simultaneous scenario as per \cite{ma-etal-2020-simulmt}. 
Moreover, the encoder self-attentions are biased using a logarithmic distance penalty \citep{Gangi2019}, leveraging the local context. Training was performed with cross entropy loss, Adam optimizer \citep{DBLP:journals/corr/KingmaB14} with a learning rate of 1e-4 with an inverse square-root scheduler and 4,000 warm-up updates. We set mini-batches of 5,000 max tokens and update the gradients every 16 mini-batches. The best checkpoint was selected based on the lowest cross entropy value on the MuST-Cinema dev set.
READ/WRITE actions of the wait-k policy are decided by means of a pre-decision module at BPE (token) level \citep{ma-etal-2020-simulmt}. The adopted pre-decision module is fixed, which triggers the decision process at every pre-defined number of frames. 
Since a frame covers 10ms of the audio, an encoder state covers 40ms due to a 4x subsampling by the initial convolutional layers. Since the average length of a word in MuST-Cinema is 270ms, we consider 7 encoder states for a READ/WRITE action, which is the default parameter used by \cite{ma-etal-2020-simulmt}, leading to a window size of 280ms. 
In order to explore the quality vs latency compromise and to study the effect of the system latency on the delay of the subtitles, we experimented with two values of $k$, resulting in wait-3 and wait-5 models. 
For comparison, we also trained one offline ST system per language.

\paragraph{Evaluation}
The evaluation focuses on two different aspects: 1)  systems' performance and 2)  display modes. 
For systems' performance, quality is evaluated with SacreBLEU \citep{post-2018-call}, which is computed on the ST output containing the subtitle breaks. The latency of the system is evaluated with Average Lagging (AL) \citep{ma-etal-2019-stacl} adapted to the ST scenario by \cite{ma-etal-2020-simulmt}. In order to test the ability of the systems to generate properly formed subtitles, we evaluate the conformity to the length constraint (Len) as the percentage of subtitles having a length between 6 and 42 characters per line \cite{ted}. The display modes are evaluated in terms of reading speed and delay, as described in Section~\ref{ssec:eval}. 

\section{Results}

\subsection{Quality, Latency and Conformity}
As far as quality is concerned (Table~\ref{tab:BLEUALresults}), the wait-3 strategy achieves low BLEU scores but there is significant improvement for wait-5, even reaching the performance of the offline system for French. These scores are in line with those reported in SimulST settings while in our setting the difficulty is exacerbated by the requirement to correctly place the subtitling breaks. In fact, the offline system, despite not being optimised for the offline mode, still performs comparatively or better than \cite{karakanta-etal-2020-42}, who reported 18.76 BLEU points for French and 11.82 for German, while the length conformity is higher by 2\%. As for latency (AL), we observe an increase between 0.2-0.6 seconds from wait-3 to wait-5, which lags behind by 2 seconds. Still, these spans are not higher than the Ear-to-Voice Span (EVS) threshold reported for intralingual respeaking (2.1 seconds) and way below the EVS for interlingual respeaking (4.1 secs) \citep{ear-voice-span}. This shows that, despite the poor quality, SimulST could have the potential of reducing the delay in interlingual live subtitling. In terms of proper subtitles, we found that our systems are capable of properly inserting the break symbols, despite the partial input they receive, since more than 90\% of the generated subtitles conform to the length constraint. This ability of our SimulST systems is indispensable for taking advantage of the predicted structure of the subtitles to experiment with display modes in blocks and lines.

\begin{table}[ht]
\centering
\begin{tabular}{l|c|c|c|c|c|c|c|c|c}
 \hline
 \multirow{2}{*}{\textbf{Model}} & \multicolumn{3}{c|}{\textbf{en-it}} & 
  \multicolumn{3}{c|}{\textbf{en-de}} & 
  \multicolumn{3}{c}{\textbf{en-fr}} \\
 \cline{2-10}
  & BLEU & AL & Len & 
  BLEU & AL & Len & 
  BLEU & AL & Len \\
  \hline
 \emph{offline} & 19.5 & - & 96\% & 14.0 & - & 96\% & 18.6 & - & 97\% \\
 wait-3 & 12.2 & 1755 & 91\% & 7.7 & 1422 & 94\% & 13.5 & 1570 & 92\% \\
 wait-5 & 15.1 & 1936 & 92\% & 11.1 & 2050 & 91\% & 18.1 & 2035 & 94\% \\
 \hline
\end{tabular}
\caption{SacreBLEU (considering \eol and \eob), Average Lagging (AL) in ms and conformity to the length constraint (Len) on three language pairs of MuST-Cinema amara.}
\label{tab:BLEUALresults}
\end{table}

\subsection{Display mode and reading speed}
When comparing the reading speed (\textit{rs}) of the three modes (Table~\ref{tab:readingspeed}), the word-for-word and block mode have the highest \textit{rs} for the wait-5 and wait-3 strategy respectively. 
The standard deviation is much higher for the word-for-word mode, which indicates a large variation in the \textit{rs}. This could be attributed to the SimulST systems' generation rate. The systems wait at the beginning of the utterance but, when the end of the input is reached, they perform greedy search and emit all remaining words at once. This rate leads to a jerky display of words, where some words remain on screen for a long time and others flush before the viewer manages to read them. However, the block mode has the lowest percentage of subtitles achieving a reading speed of max 21 cps. The problem of this mode is that each block remains on screen until the next \eob is generated, which corresponds to the duration of the following block. For example, if a block of two lines with 40 characters each is followed by a block of one line of 25 characters, the first block would have a short time to be displayed, resulting in a high \textit{rs}, and vice versa. One future direction would be to adjust the time of each block to better accommodate its reading speed, however, in initial experiments we found that this approach led to excessively high delay. Scrolling lines, our proposed method, achieves by far the lowest mean \textit{rs}, with all models scoring below the 21 cps threshold. The same result is shown for the percentage of conforming subtitles, where conformity to reading speed reaches $\sim$80\%. It is worth noting that \textit{rs} increases from wait-3 to wait-5 for the block and line modes for en$\rightarrow$de, contrary to the other languages. This correlates with the lower percentage of length conformity (94\% for wait-3 to 91\% for wait-5) and shows the importance of correctly predicting the position of the breaks for the success of the display methods relying on these breaks.

As for delay, the word-for-word mode has the lowest delay, which corresponds to the system's lagging. The block mode has the highest delay, while our proposed method manages to reduce the delay by 0.6 seconds on average compared to the display in blocks, remaining close to a 4-second EVS. 
Our results are validated by the inversely proportional relationship between \textit{rs} and delay. Scrolling lines, our proposed method, seems to achieve a fair compromise between a comfortable reading speed and an acceptable delay, while combining the benefits of the presence of the word on the right, less dynamic text and the preservation of the block structure which is familiar to most viewers.

\begin{table}[ht]
\centering
\begin{tabular}{l|l|c|c|c|c|c|c}
 \hline
 \multirow{6}{*}{\rotatebox[origin=c]{90}{\textbf{en-it}}} & display & \multicolumn{3}{c|}{\textit{wait-3}} & 
  \multicolumn{3}{c}{\textit{wait-5}} \\
 \cline{3-8} 
  & mode & rs & $\leq21$ cps & delay &
  rs & $\leq21$ cps & delay \\ 
 \cline{2-8} 
 &word & 53.5 $\pm$ 9.9 & 61\% & 1755 & 40.1 $\pm$ 8.0 & 70\% & 1936 \\
 &block & 53.4 $\pm$ 9.1 & 38\% & 4690 & 36.5 $\pm$ 7.0 & 62\% & 5004 \\
 &line & 17.6 $\pm$ 5.2 & 79\% & 4092 & 14.4 $\pm$ 4.1 & 85\% & 4461 \\
 \hline
  \multirow{6}{*}{\rotatebox[origin=c]{90}{\textbf{en-de}}} & display & \multicolumn{3}{c|}{\textit{wait-3}} & 
  \multicolumn{3}{c}{\textit{wait-5}} \\
 \cline{3-8} 
 & mode & rs & $\leq21$ cps & delay &
  rs & $\leq21$ cps & delay \\
 \cline{2-8} 
 &word & 29.1 $\pm$ 7.2 & 70\% & 1422 & 58.4 $\pm$ 10.5 & 63\% & 2050 \\
 &block & 33.3 $\pm$ 6.2 & 37\% & 4772 & 52.6 $\pm$ 9.2 & 56\% & 4503\\
 &line & 12.4 $\pm$ 3.7 & 85\% & 4090 & 19.9 $\pm$ 5.4 & 78\% & 3894\\
  \hline
  \multirow{6}{*}{\rotatebox[origin=c]{90}{\textbf{en-fr}}} & display & \multicolumn{3}{c|}{\textit{wait-3}} & 
  \multicolumn{3}{c}{\textit{wait-5}} \\
 \cline{3-8} 
  &mode & rs & $\leq21$ cps & delay &
  rs & $\leq21$ cps & delay \\
 \cline{2-8} 
 &word & 39.7 $\pm$ 7.9 & 55\% & 1570 & 53.4 $\pm$ 9.0 & 57\% & 2035\\
 &block & 43.8 $\pm$ 7.6 & 37\% & 4872 & 46.1 $\pm$ 8.0 & 56\% & 5273 \\
 &line & 15.4 $\pm$ 4.2 & 78\% & 4217 & 18.4 $\pm$ 4.8 & 78\% & 4708\\
 \hline
\end{tabular}
\caption{Reading speed (\textit{rs}) mean and standard deviation  in characters per second (cps), percentage of subtitles with a \textit{rs} of max. 21 cps and display delay (in ms) on three language pairs of MuST-Cinema amara.}
\label{tab:readingspeed}
\end{table}


\section{Conclusions}
\label{sec:conclusions}
In this work we adapted SimulST systems for the task of live subtitling, by forcing the systems to generate subtitle breaks. We showed that SimulST systems are able to generate properly-formed subtitles. Given this finding, we moved on to explore display strategies alternative to the word-for-word display, the established display mode in SimulST. Word-for-word display is sub-optimal for readability and comprehension \citep{Romero-fresco-2010-Standingonquicksand}. For automatically generated live subtitles, we found that it leads to an extremely variable reading speed, with some words lagging on the screen while the words towards the end of the utterance flushing through the screen. On the other hand, the display in blocks, which is the traditional mode for displaying offline subtitles, leads to a large delay and improves the reading speed only for SimulST systems with a higher latency. Our proposed display method, scrolling lines, is the only one achieving a comfortable mean reading speed below 21 cps, with around 80\% of the subtitles having acceptable reading speed, while the delay remains along the 4-second threshold. 

As for the feasibility of SimulST for live subtitling, there is still a long way to go in several directions. From a technical point of view, the principal issue is still the poor translation quality, which could benefit from advancements in tailored architectures. Evaluation marks progress in any field, but we still lack robust evaluation methodologies, taking into account all dimensions of the target medium, both quality and readability. 
Lastly, scholarly work is needed around user perception studies in automatic subtitling and live interlingual subtitling. Technology is moving faster than research and user studies are key to ensure our implementational efforts are moving in the right direction. We hope our work has set the ball rolling for further research in automatising live interlingual subtitling.


\small

\bibliographystyle{apalike}
\bibliography{mtsummit2021}

\end{document}